\documentclass[sigconf]{acmart}
\AtBeginDocument{%
  }
\setcopyright{acmlicensed}

\copyrightyear{2025}
\acmYear{2025}
\setcopyright{acmlicensed}\acmConference[WWW Companion '25]{Companion Proceedings of the ACM Web Conference 2025}{April 28-May 2, 2025}{Sydney, NSW, Australia}
\acmBooktitle{Companion Proceedings of the ACM Web Conference 2025 (WWW Companion '25), April 28-May 2, 2025, Sydney, NSW, Australia}
\acmDOI{10.1145/3701716.3715189}
\acmISBN{979-8-4007-1331-6/2025/04}

\begin{document}

\title{OneKE: A Dockerized Schema-Guided LLM Agent-based Knowledge Extraction System}

\author{Yujie Luo}
\email{luo.yj@zju.edu.cn}
\affiliation{%
  \institution{Zhejiang University}
  \institution{ZJU-Ant Group Joint Research Center for Knowledge Graphs}
  \city{Hangzhou}
  \country{China}
}

\author{Xiangyuan Ru}
\author{Kangwei Liu}
\affiliation{%
  \institution{Zhejiang University}
  \institution{ZJU-Ant Group Joint Research Center for Knowledge Graphs}
  \city{Hangzhou}
  \country{China}
}

\author{Lin Yuan}
\author{Mengshu Sun}
\affiliation{%
  \institution{ZJU-Ant Group Joint Research Center for Knowledge Graphs}
  \institution{Ant Group}
  \city{Hangzhou}
  \country{China}
}

\author{Ningyu Zhang*}
\affiliation{%
  \institution{Zhejiang University}
  \institution{ZJU-Ant Group Joint Research Center for Knowledge Graphs}
  \city{Hangzhou}
  \country{China}
}

\author{Lei Liang}
\author{Zhiqiang Zhang}
\affiliation{%
  \institution{ZJU-Ant Group Joint Research Center for Knowledge Graphs}
  \institution{Ant Group}
  \city{Hangzhou}
  \country{China}
}

\author{Jun Zhou*}
\affiliation{%
  \institution{ZJU-Ant Group Joint Research Center for Knowledge Graphs}
  \institution{Ant Group}
  \city{Hangzhou}
  \country{China}
}

\author{Lanning Wei}
\author{Da Zheng}
\affiliation{%
  \institution{ZJU-Ant Group Joint Research Center for Knowledge Graphs}
  \institution{Ant Group}
  \city{Hangzhou}
  \country{China}
}

\author{Haofen Wang}
\affiliation{%
  \institution{Tongji University}
  \city{Shanghai}
  \country{China}
}

\author{Huajun Chen*}
\affiliation{%
  \institution{Zhejiang University}
  \institution{ZJU-Ant Group Joint Research Center for Knowledge Graphs}
  \city{Hangzhou}
  \country{China}
}

\renewcommand{\shortauthors}{Luo et al.}

\begin{abstract}
We introduce OneKE, a dockerized schema-guided knowledge extraction system, which can extract knowledge from the Web and raw PDF Books, and support various domains (science, news, etc.). Specifically, we design OneKE with multiple agents and a configure knowledge base. Different agents perform their respective roles, enabling support for various extraction scenarios. The configure knowledge base facilitates schema configuration, error case debugging and correction, further improving the performance. Empirical evaluations on benchmark datasets demonstrate OneKE's efficacy, while case studies further elucidate its adaptability to diverse tasks across multiple domains, highlighting its potential for broad applications. We have open-sourced the Code\footnote{\url{https://github.com/zjunlp/OneKE}} and released a Video\footnote{\url{http://oneke.openkg.cn/demo.mp4}}.
\end{abstract}



\maketitle

\section{Introduction}

Knowledge extraction--obtaining knowledge from data, is a critical component for a wide range of practical systems such as Knowledge Graph (KG) construction \cite{chen2022knowprompt}, Retrieval Augmentation (RAG) \cite{gao2023retrieval}, and domain-specific applications like scientific discovery \cite{dagdelen2024structured} and intelligence analysis \cite{sun2023cyber}. 
The last decades have witnessed the development of various knowledge extraction systems \cite{li2023evaluating,wei2024chatie,xu2024large}. 
In particular, with the emergence of Large Language Models (LLMs), new works such as InstructUIE \cite{DBLP:journals/corr/abs-2304-08085}, iText2KG \cite{DBLP:journals/corr/abs-2409-03284} and AgentRE \cite{DBLP:conf/cikm/ShiJQY24} have been continuously emerged.
However, previous approaches still struggle to effectively extract information from raw data following complex schemas and face challenges in debugging and correcting errors when they occur.

\begin{figure*}[ht]
    \centering
    \includegraphics[width=\linewidth]{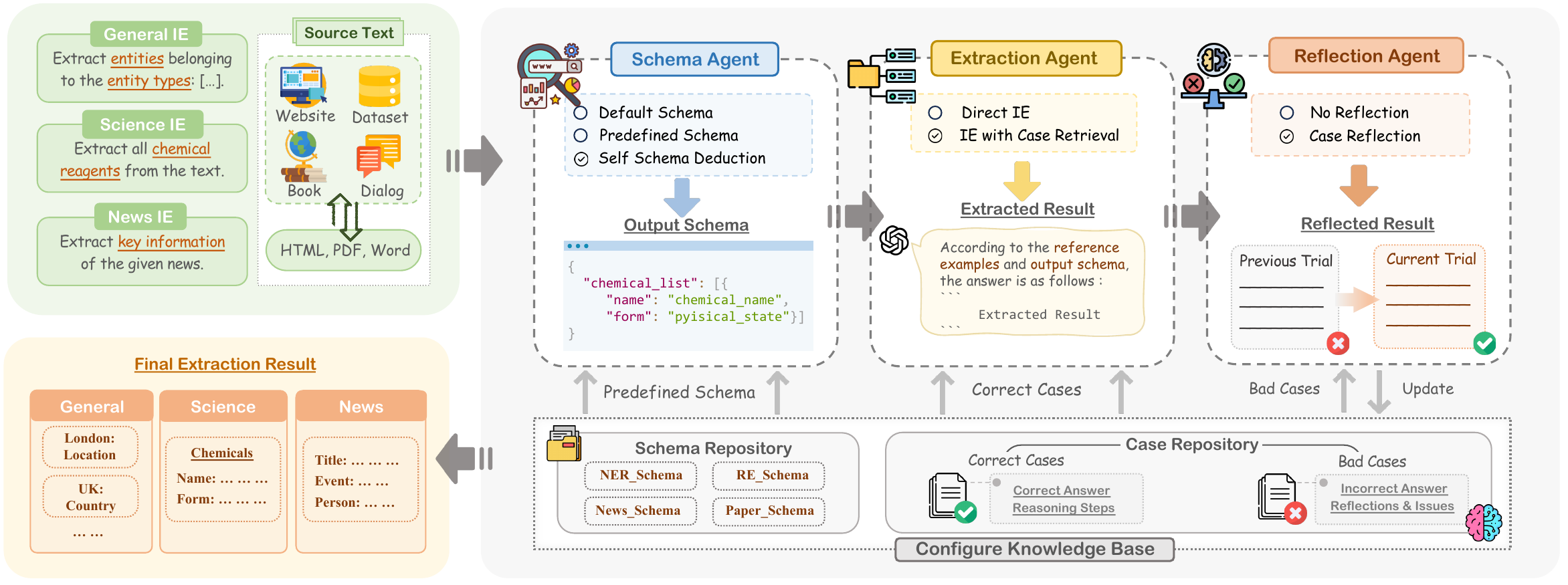}
    \caption[width=\linewidth]
    {The overview of the OneKE system, supporting various domains (science, news, etc.) and data (Web HTML, PDF, etc.).}
    \label{fig:main}
\end{figure*}

Note that previous efforts have primarily focused on the capabilities of individual models while neglecting the design of a comprehensive system to address the knowledge extraction task. 
To this end, we introduce OneKE, a dockerized schema-guided knowledge extraction system.
We adopt a multi-agent design with a configure knowledge base to provide knowledge extraction support for various scenarios and error debugging, aiming to meet the practical needs of users as much as possible.
Following \cite{DBLP:conf/cikm/ShiJQY24}, we design three agents: \textbf{Schema Agent} for schema analysis with various data types, \textbf{Extraction Agent} for extracting knowledge with various LLMs, and \textbf{Reflection Agent} to debug and handle erroneous cases.
Based on this design, OneKE can efficiently process source texts of varying lengths and formats, such as HTML and PDF, and demonstrates a robust capability to adapt to diverse task configurations, yielding a comprehensive range of output schemas tailored to specific requirements.

We evaluate OneKE using two benchmark datasets for Named Entity Recognition (NER) and Relation Extraction (RE), demonstrating the effectiveness of our framework. 
Furthermore, to explore the versatility of OneKE in practical applications, we conduct case analyses on specific extraction tasks.
These scenarios encompass extracting structured information from \textbf{the Web news articles} and \textbf{raw PDF book chapters}, highlighting OneKE's capability to manage diverse data formats and varying task contexts effectively.
This flexible framework, which operates without the necessity of fine-tuning, is adept at swift adaptation to forthcoming LLMs, thereby amplifying their capabilities and elevating their overall efficacy.

\section{Design and Implementation}

OneKE is thoughtfully designed to address the complexities and challenges inherent in knowledge extraction.
As shown in Figure \ref{fig:main}, the framework is guided by several key considerations that enhance its functionality and adaptability in real-world scenario:

\textbf{(1) Adaptability to real-world data.} 
Real-world information extraction tasks often handle raw data, like HTML, PDF, etc. 
Based on this, the OneKE framework supports a variety of data types rather than pure text.
We also reserve a user-defined interface to support new data types in the future.

\textbf{(2) Generalization for complex schemas.} 
Practical knowledge extraction scenarios should handle diverse and complex schemas, or even no schema. 
Thus, we design the OneKE-specific Schema Agent to support both pre-defined schemas and self-schema deduction using LLMs. 
OneKE also supports various LLMs, including LLaMA, Qwen and ChatGLM, as well as proprietary models like GPT-4, enabling effective knowledge extraction.

\textbf{(3) Debugging and fixing errors.} 
Most previous works require retraining the model when encountering error cases. 
In contrast, we integrate Case Repository into OneKE to equip the model with reflective and error-correcting capabilities, enabling its continuous improvement in knowledge extraction tasks.

\subsection{Schema Agent}

To support various task settings and data types, we develop the Schema Agent based on LLMs to generate the corresponding output schema for each task.
The primary goal is to preprocess the data, standardize its format and schema, and prepare it for the subsequent information extraction step.
To support various real-world data, we utilize the \textit{document\_loaders} module provided by the Langchain, 
to preprocess the data and perform chunking on long texts.
Users can also define new data types, and add custom preprocessing methods.
Next, if the user has defined a schema, given a task description and raw data input, the Schema Agent will select the appropriate pre-defined schema from the schema repository in the configure knowledge base.
If the user does not provide a schema, the model will generate a unified output schema based on the user's instructions, such as ``\emph{Extract characters and background setting}''. 
Users can customize schemas using simple text by updating the schema repository in the Configure Knowledge Base.

\subsection{Extraction Agent} 
\label{sec:second_section}
Upon receiving the unified output schema from the Schema Agent, we design the Extraction Agent to utilize LLMs for extracting knowledge, thereby generating the preliminary extraction results.
Specifically, this module supports a variety of models, including locally deployed open-source models such as LLaMA, Qwen, and ChatGLM, as well as API services like OpenAI and DeepSeek. 
To enhance performance, the Extraction Agent learns from similar cases and applies this knowledge to the extraction process.
Relevant cases are retrieved from the Case Repository using semantic similarity combined with string matching (via the all-MiniLM-L6-v2 model and the FuzzyWuzzy package, we set Top two as default).
These cases are then incorporated as few-shot examples into the original context to form the prompt, after which the LLM is called to obtain the extraction results.
After the above steps, we obtain the preliminary extraction results; however, errors often occur. 
To address this, we design a reflection agent to debug and fix these errors.
We use self-consistency to filter out the cases where the model is uncertain and pass these cases to the Reflection Agent.

\subsection{Reflection Agent}
\label{third:third_section}

To enable debugging and error correction, allowing OneKE to learn from past mistakes, we follow \cite{DBLP:conf/cikm/ShiJQY24} to design the Reflection Agent to facilitate reflection and optimization.
By leveraging prior knowledge, the agent refines and improves the initial output from the previous module, ultimately producing the final extraction result.
Concretely, the agent leverages external Case Repository, specifically relevant bad cases, to facilitate reflection and correction. 
In a manner similar to the retrieval approach discussed in Section \ref{sec:second_section}, the Reflection Agent fetches bad cases that are most relevant to the current task and extraction text. 
These relevant bad cases, along with their associated reflective analyses, are subsequently incorporated into the LLM.
In this way, the agent can effectively learn from past mistakes, enabling its error correction capabilities to generate accurate answers.

\subsection{Configure Knowledge Base}
The Configure Knowledge Base provides essential information for the three agents, including manually defined schemas for various tasks, and historical extraction cases, enhancing the performance and error correction capabilities:

\noindent \textbf{Schema Repository.} 
In the OneKE system, the Schema Repository provides the pre-defined schema for the Schema Agent, supporting the subsequent extraction process. 
Specifically, the Schema Repository includes pre-defined output schemas for NER, RE, and EE tasks, along with templates for various data scenarios such as \textit{scientific academic papers} and \textit{news reports}.
The schemas in the Schema Repository are structured as Pydantic objects, enabling seamless serialization into JSON format for the Extraction Agent. 
Moreover, this structure allows users to customize new schemas within the repository, thus enhancing adaptability and extensibility.

\noindent \textbf{Case Repository.}
To enable debugging and error correction in OneKE, we design the Case Repository, which primarily stores traces of past knowledge extraction cases. 
This repository supports the Extraction Agent in performing extractions and assists the Reflection Agent in reflecting on and correcting errors.
Specifically, knowledge extraction cases stored in Case Repository can be divided into two categories: Correct Cases and Bad Cases.
Correct Cases provide the Extraction Agent with reasoning steps of successful extraction, while Bad Cases offer the Reflection Agent warnings about avoidable mistakes.
The Case Repository will be automatically updated once a knowledge extraction task is completed.
Concretely, this module first generates reasoning steps derived from the correct answer, storing both the correct answer and its reasoning steps in the Correct Case Repository to enhance task understanding. 
Additionally, the agent compares its answer with the correct one and reflects on its original response to identify potential issues. 
It then stores the original answer along with the corresponding reflections in the Bad Case Repository for future reference.

\begin{figure}[ht] 
    \centering
    \includegraphics[width=0.49\linewidth]{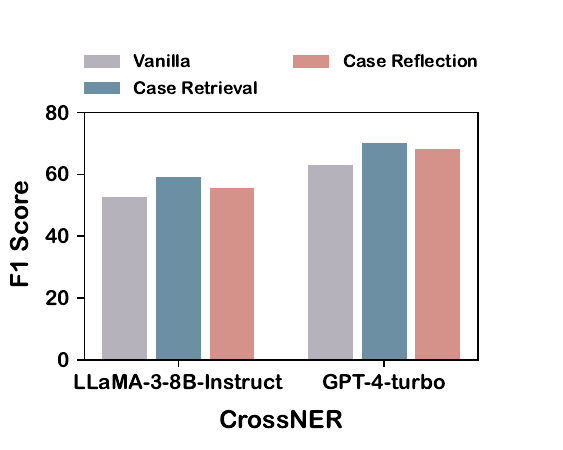} 
    \includegraphics[width=0.49\linewidth]{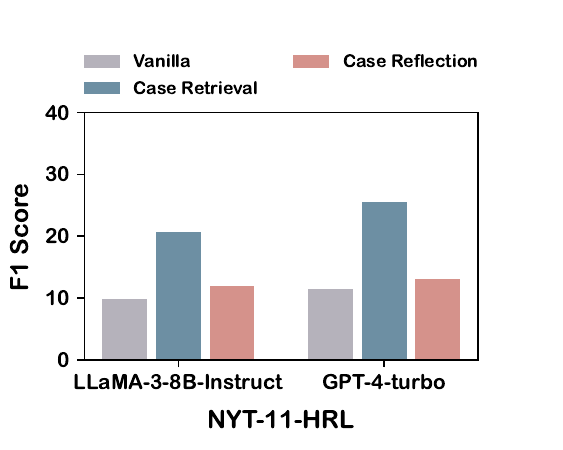}
    \caption{Performance of different components in OneKE.}
    \label{fig:main_results}
\end{figure}

\section{Evaluation}
 
\begin{figure*}[!t]
    \centering
    \includegraphics[width=0.88\linewidth]{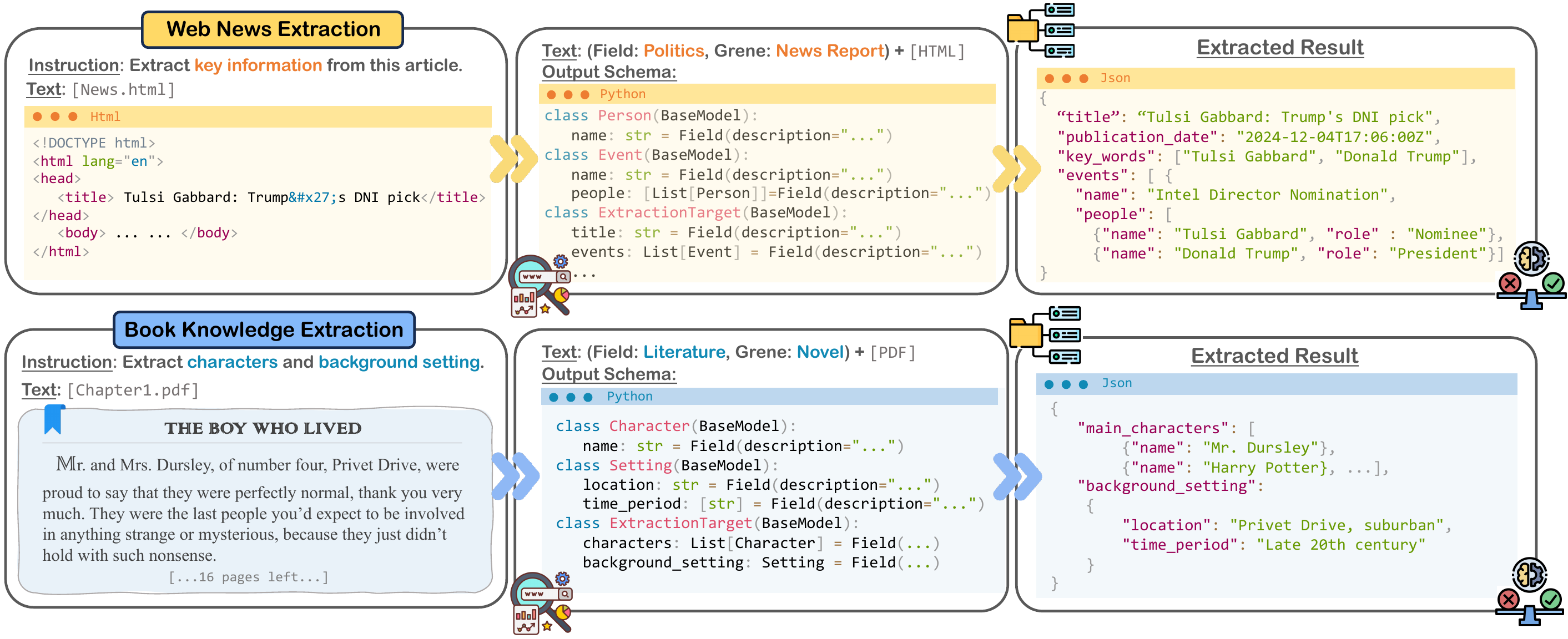}
    \caption{Using OneKE on Web News Extraction and Book Knowledge Extraction.}
    \label{fig:case}
\end{figure*}

\noindent \textbf{Experimental Settings. } 
We evaluate OneKE on the CrossNER and NYT-11-HRL datasets. 
CrossNER is a cross-domain NER dataset, while NYT-11-HRL focuses on the RE task within the news domain. 
The performance of OneKE is evaluated on both the LLaMA-3-8B-Instruct and GPT-4-turbo models.

\noindent \textbf{Main Results.} 
As depicted in Figure \ref{fig:main_results}, the various methods employed in OneKE confer performance enhancements across both NER and RE. 
Notably, the Case Retrieval method of the Extraction Agent achieves the most significant improvements. 
Through analysis, we observe that the agent effectively applies the reasoning paths in the provided samples, thereby facilitating accurate extraction.
Additionally, by comparing the two different tasks, we observe that the aforementioned Case Retrieval method is more effective for the more challenging RE task, as the intermediate reasoning steps are essential in such complex scenarios. 
The Case Reflection primarily emphasizes the model's ability to recognize known errors and its capacity for transfer learning regarding those errors, leading to similar improvements across both tasks.
We provide case studies of specific application scenarios in the following Section.

\section{Application}
In practical applications, the OneKE framework supports diverse data formats (HTML, PDF, Word), accommodating both short and long contexts for seamless integration into various downstream applications.
We provide case analyses in the following two representative extraction scenarios.

\noindent \textbf{Web News Extraction.} 
In the news domain, OneKE enhances the knowledge extraction of Web news content, thereby facilitating downstream tasks such as effective sentiment monitoring, proactive risk management, as well as a variety of additional applications.
As illustrated in Figure \ref{fig:case}, the extraction task starts with \textit{Extracting key information from news articles} on a randomly selected \textbf{raw HTML page} from the web, aiming to identify the overall nature of the news and obtain structured key insights.
After parsing the HTML-formatted text, the Schema Agent first identifies its domain and genre as a Politics News Report, offering crucial guidance. 
Utilizing this metadata, the Schema Agent generates a structured Output Schema in code format that effectively captures the \textit{key information} of the news.
Once the Output Schema has been serialized into a JSON format description, the Extraction Agent and the Reflection Agent collaborate to undertake subsequent extraction and reflective optimization tasks. 
This cooperative effort of the agents culminates in a JSON output that captures the key information and structure of the news report.

\noindent \textbf{Book Knowledge Extraction.} 
Another application of OneKE lies in extracting structured knowledge from extensive corpora, including books, documents, or manuals.
Specifically, we use the first chapter of the \textit{Harry Potter series} (with a total length of 17 pages in PDF format) as the target text.
The extraction focuses on the \textit{main characters} and the \textit{background setting} within this chapter.
As shown in Figure \ref{fig:case}, the generated Output Schema accurately identified the two target extraction objects: \textit{``main\_characters''} and \textit{``background\_setting''}. 
Subsequently, with the collaboration of the Extraction Agent and the Reflection Agent, OneKE successfully extracted relevant information.

\section{Conclusion and Future Work}
In this paper, we introduce OneKE, a Dockerized Schema-Guided LLM Agent-based Knowledge Extraction System. 
OneKE is designed for flexible application across a spectrum of extraction tasks in real-world scenarios.
It can handle source texts of varying lengths and formats (such as HTML and PDF) while demonstrating the capability to adapt to diverse task configurations, generating a broad range of output schemas tailored to specific requirements.
Moreover, the integration of a self-reflection mechanism enables iterative improvement informed by external feedback, thereby improving both accuracy and adaptability.

\noindent \textbf{Long-term Maintenance.}  We will maintain OneKE over the long term, adding new features and fixing bugs.
To advance document data extraction and comprehension, we plan to develop methodologies for the integration and analysis of diverse chart types and content.
We hope OneKE can serve as a helpful tool for researchers and engineers engaging in knowledge extraction with LLMs.

\begin{acks}
This work was supported by the National Natural Science Foundation of China (No. 62206246, No. U23B2057, No. 62176185), the Fundamental Research Funds for the Central Universities (226-2023-00138), Yongjiang Talent Introduction Programme (2021A-156-G), CIPSC-SMP-Zhipu Large Model Cross-Disciplinary Fund, Ningbo Science and Technology Special Projects under Grant No. 2023Z212.
This work was supported by Ant Group and Zhejiang University - Ant Group Joint Laboratory of Knowledge Graph.
\end{acks}

\bibliographystyle{ACM-Reference-Format}
\bibliography{sample-base}


\end{document}